\documentclass[a4paper]{article}

\usepackage{arxiv}

\usepackage[utf8]{inputenc}       % allow utf-8 input
\usepackage[T1]{fontenc}          % use 8-bit T1 fonts
\usepackage{hyperref}             % hyperlinks
\usepackage{url}                  % simple URL typesetting
\usepackage{booktabs}             % professional-quality tables
\usepackage{amsfonts}             % blackboard math symbols
\usepackage{amsmath}              % popular math symbols
\usepackage{nicefrac}             % compact symbols for 1/2, etc.
\usepackage{microtype}            % microtypography
\usepackage[capitalise]{cleveref} % smart cross-referencing
\usepackage{lipsum}               % Can be removed after putting your text content
\usepackage{graphicx}
\usepackage{natbib}
\usepackage{doi}
\usepackage[nolist]{acronym}
\usepackage[dvipsnames,table]{xcolor}
\usepackage[most]{tcolorbox}    % For creating colored boxes
\usepackage{siunitx}
\usepackage{soul}
\usepackage{rotating}

\newcommand{\hlc}[2][yellow]{{%
    \colorlet{foo}{#1}%
    \sethlcolor{foo}\hl{#2}}%
}

\title{Large Language Models for Human-Machine Collaborative Particle Accelerator Tuning through Natural Language}
\date{14 May 2024}
\newif\ifuniqueAffiliation
\ifuniqueAffiliation % Standard variant of author block
\author{ \href{https://orcid.org/0000-0000-0000-0000}{\includegraphics[scale=0.06]{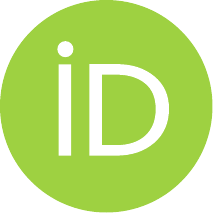}\hspace{1mm}David S.~Hippocampus}\thanks{Use footnote for providing further
		information about author (webpage, alternative
		address)---\emph{not} for acknowledging funding agencies.} \\
	Department of Computer Science\\
	Cranberry-Lemon University\\
	Pittsburgh, PA 15213 \\
	\texttt{hippo@cs.cranberry-lemon.edu} \\
	\And
	\href{https://orcid.org/0000-0000-0000-0000}{\includegraphics[scale=0.06]{orcid.pdf}\hspace{1mm}Elias D.~Striatum} \\
	Department of Electrical Engineering\\
	Mount-Sheikh University\\
	Santa Narimana, Levand \\
	\texttt{stariate@ee.mount-sheikh.edu} \\
}
\else
\usepackage{authblk}

\newbox{\orcid}\sbox{\orcid}{\includegraphics[scale=0.06]{orcid.pdf}} 
\author[1]{%
	\href{https://orcid.org/0000-0003-3445-0678}{\usebox{\orcid}\hspace{1mm}Jan Kaiser\thanks{\texttt{jan.kaiser@desy.de}}}%
}
\author[1,2]{%
	\href{https://orcid.org/0000-0003-3282-3135}{\usebox{\orcid}\hspace{1mm}Annika Eichler\thanks{\texttt{annika.eichler@desy.de}}}%
}
\author[3]{%
    \href{https://orcid.org/0000-0001-8590-9827}{\usebox{\orcid}\hspace{1mm}Anne Lauscher\thanks{\texttt{anne.lauscher@uni-hamburg.de}}}%
}
\affil[1]{Deutsches Elektronen-Synchrotron DESY, Germany}
\affil[2]{Hamburg University of Technology, 21073 Hamburg, Germany}
\affil[2]{Universität Hamburg, Germany}
\fi

\renewcommand{\shorttitle}{Large Language Models for Particle Accelerator Tuning}

\hypersetup{
pdftitle={A template for the arxiv style},
pdfsubject={q-bio.NC, q-bio.QM},
pdfauthor={David S.~Hippocampus, Elias D.~Striatum},
pdfkeywords={First keyword, Second keyword, More},
}

\newtcolorbox{promptbox}{%
fontupper=\small\ttfamily, breakable, halign=flush left, enhanced,% 
}

\begin{acronym}
    \acro{AGI}{artificial general intelligence}
    \acro{AI}{artificial intelligence}
    \acro{ANN}{artificial neural networks}
    \acro{ARES}{accelerator research experiment at SINBAD}
    \acro{BPM}{beam position monitor}
    \acro{BO}{Bayesian optimisation}
    \acro{CI}{continuous integration}
    \acro{CoT}{Chain-of-Thought}
    \acro{CPU}{central processing unit}
    \acro{CUDA}{Compute Unified Device Architecture}
    \acro{EI}{expected improvement}
    \acro{EA}{Experimental Area}
    \acro{ES}{extremum seeking}
    \acro{FDF}{focus-defocus-focus}
    \acro{FEL}{free-electron laser}
    \acro{GAIA}{general AI assistant for intelligent accelerator operations}
    \acro{GAN}{generative adversarial network}
    \acro{GP}{Gaussian process}
    \acro{GPGPU}{general purpose GPU}
    \acro{GPU}{graphics processing unit}
    \acro{HPC}{high-performance computing}
    \acro{LLM}{large language model}
    \acro{MAE}{mean absolute error}
    \acro{ML}{machine learning}
    \acro{MLP}{multilayer perceptron}
    \acro{MMLU}{Massive Multitask Language Understanding}
    \acro{MPS}{Metal Performance Shaders}
    \acro{MSE}{mean squared error}
    \acro{NLP}{natural language processing}
    \acro{NN}{neural network}
    \acro{PPO}{Proximal Policy Optimisation}
    \acro{RCDS}{robust conjugate direction search}
    \acro{ReLU}{rectified linear unit}
    \acro{RMSE}{root mean squared error}
    \acro{RL}{reinforcement learning}
    \acro{RLO}{reinforcement learning-trained optimisation}
    \acro{SGD}{stochastic gradient descent}
    \acro{SOC}{system on a chip}
    \acro{TD3}{Twin Delayed DDPG}
    \acro{UCB}{upper confidence bound}
\end{acronym}

\begin{document}
\maketitle

% TODO Copy right notice

\begin{abstract}
    Autonomous tuning of particle accelerators is an active and challenging field of research with the goal of enabling novel accelerator technologies cutting-edge high-impact applications, such as physics discovery, cancer research and material sciences.
    A key challenge with autonomous accelerator tuning remains that the most capable algorithms require an expert in optimisation, machine learning or a similar field to implement the algorithm for every new tuning task.
    In this work, we propose the use of \acfp{LLM} to tune particle accelerators.
    We demonstrate on a proof-of-principle example the ability of \acp{LLM} to successfully and autonomously tune a particle accelerator subsystem based on nothing more than a natural language prompt from the operator, and compare the performance of our \ac{LLM}-based solution to state-of-the-art optimisation algorithms, such as \acf{BO} and \acf{RLO}.
    In doing so, we also show how \acp{LLM} can perform numerical optimisation of a highly non-linear real-world objective function.
    Ultimately, this work represents yet another complex task that \acp{LLM} are capable of solving and promises to help accelerate the deployment of autonomous tuning algorithms to the day-to-day operations of particle accelerators.
\end{abstract}

% keywords can be removed

\section{Introduction}

Particle accelerators are sophisticated machines designed to accelerate subatomic particles, such as electrons and protons, to extremely high speeds, often close to the speed of light. These devices play a crucial role in a variety of applications, ranging from fundamental research in physics to practical uses in medicine, such as cancer therapy, and material science. As the demands from these diverse applications grow, there is an increasing need for  advanced tuning and control methods to manage the complex dynamics of particle acceleration.
Despite this, as a result of its complexity, the tuning of particle accelerators is to this day often done manually by experienced human operators.
In this context, the emergence of autonomous tuning methods represents a significant advancement. By leveraging methods from the fields of numerical optimisation and machine learning~\citep{emery2003use,roussel2023bayesian,kaiser2022learningbased}, autonomous systems promise to speed up accelerator tuning procedures, reducing costs and minimising downtime, while also enabling novel operating modes for state-of-the art measurements. Moreover, such methods enable a paradigm shift from actuator-driven accelerator operation, where human operators control actuator settings to achieve good measurement conditions, to specification-driven operation, where human operators determine the best conditions for experiments and autonomous agents ensure that these conditions are achieved.
As such, autonomous particle accelerator tuning methods promise to not only improve the performance of accelerators on existing applications but also open up new possibilities in scientific research and industrial applications, marking a transformative step in the field of particle acceleration.

However, implementing state-of-the-art accelerator tuning methods on new tuning tasks requires experts in two separate domains -- accelerator physics and optimisation -- as well as significant engineering effort to solve problems ranging from algorithm selection to objective function formulation.
These challenges have so far slowed the adoption of advanced autonomous tuning algorithms to day-to-day accelerator operations.

In recent developments, \acp{LLM}, such as \textit{GPT 4}~\citep{openai2023gpt4} and \textit{Llama 2}~\citep{touvron2023llama}, have been demonstrated to be capable of solving complex tasks when prompted through natural language~\citep{brown2020language,openai2023gpt4,oulianov2024evaluate}.
The question arises whether \acp{LLM} can directly perform particle accelerator tuning, when prompted by an accelerator expert describing the tuning goal.
If capable, this would provide a more natural way of controlling autonomous tuning solutions through natural language, potentially enabling a more straightforward deployment of autonomous particle accelerator tuning solutions, and removing the requirement for optimisation algorithm-specific expertise.
Moreover, the ability of \acp{LLM} to explain their reasoning~\citep{wei2023chainofthought} could provide valuable insights into the complex dynamics of particle accelerators, potentially aiding human operators in understanding the tuning process.
Lastly, the successful application of \acp{LLM} to particle accelerator tuning would also demonstrate the ability of \acp{LLM} to solve (multi-objective) numerical optimisation problems, possibly opening up new avenues for the application of \acp{LLM} to optimisation tasks beyond particle accelerators.

In this work, we introduce a novel approach to using \acp{LLM} for autonomous tuning of a particle accelerator.
We answer whether current state-of-the-art \acp{LLM} are in fact capable of solving particle accelerator tuning tasks and evaluate our \ac{LLM}-based approach against the current state of the art in accelerator tuning using \ac{RLO} and \ac{BO}.

To this end, we review related work in \cref{sec:related_work}, before introducing our approach for autonomous tuning of a particle accelerator using \acp{LLM} and our prompt design in \cref{sec:tuning_particle_accelerators_with_llms}.
In \cref{sec:evaluation}, we evaluate the developed solution, comparing \num{14} state-of-the-art \ac{LLM} models against each other; against \num{3} state-of-the-art accelerator tuning solutions, \ac{RLO}, \ac{BO} and \ac{ES}; as well as against two baselines, random search and doing nothing.
Our findings indicate that \acp{LLM} are capable of tuning particle accelerators, but do not yet achieve performance competitive with the state of the art.
We conclude this paper and discuss opportunities for future applications of \acp{LLM} in the operation of particle accelerator facilities in \cref{sec:conclusion_and_outlook}.

\section{Related Work}\label{sec:related_work}

Initial efforts towards autonomous accelerator tuning have investigated numerical methods such as Nelder-Mead simplex~\citep{emery2003use,shang2005parallel,huang2018robust}, \acf{RCDS}~\citep{huang2013algorithm,olsson2018online,zhang2022optimization}, \acf{ES}~\citep{scheinker2022extremum} and genetic algorithms~\citep{bergan2019online}. These methods have since found adaptation in the day-to-day tuning of particle accelerator facilities~\citep{tomin2016progress,zhang2021badger,zhang2022badger}.
More recently, advanced methods like \acf{BO} have found increased interest in the accelerator community~\citep{roussel2023bayesian}
% ~\citep{duris2020bayesian,hanuka2019online,jalas2021bayesian,roussel2021multiobjective,xu2023bayesian,mcintire2016bayesian}
for their ability to utilise machine learning to learn a probabilistic surrogate model of the underlying objective function, enabling more sample-efficient tuning of high-dimensional and increasingly complex accelerator systems. Efforts are currently under way to lower the barrier of entry to these methods and increase their adoption in day-to-day accelerator operations~\citep{roussel2023xopt}.
Moreover, the accelerator community is looking increasingly to machine learning methods to aid with the challenges of accelerator tuning~\citep{edelen2018opportunities}. In particular, \ac{RL} has found adoption to accelerator control tasks~\citep{boltz2020feedback,stjohn2021realtime}. \Ac{RL} has also been successfully applied to so-called \acf{RLO}, where \ac{NN} policies are trained through \textit{optimiser learning}~\citep{andrychowicz2016learning,li2017learning,li2017learning2,chen2022learning} to be capable of sample-efficient accelerator tuning~\citep{kain2020sampleefficient,pang2020autonomous,kaiser2022learningbased,velotti2023towards}.

Most recently, \acfp{LLM} have had a highly visible impact on the field of \ac{AI} and \ac{ML}.
Usually based on the transformer \ac{NN} architecture, first introduced in \cite{vaswani2017attention}, these models are trained to perform text completion, such that they develop text understanding and text generation capabilities, which can be exploited to create chatbots.
As such, state-of-the-art \acp{LLM} like \textit{GPT 4}~\citep{openai2023gpt4} have been demonstrated to have impressive capabilities, such as text summarisation, but also the ability to solve more complex tasks like coding and general problem solving.
The field of \acp{LLM} is moving fast and seeing significant investments.
Despite their high training cost, many of these models have been released in a short time frame, both commercial and closed in nature, such as \textit{GPT 4}~\citep{openai2023gpt4}, \textit{Gemini}~\citep{geminiteam2023gemini} and \textit{Claude}~\citep{claude2023anthropic}, but also numerous open-source (or more specifically open-weights) models, such as \textit{Llama} (2)~\citep{touvron2023llama}, \textit{Orca} (2)~\citep{mukherjee2023orca,mitra2023orca}, \textit{Starling-LM}~\citep{zhu2023starling} and \textit{Mistral} / \textit{Mixtral}~\citep{jiang2023mistral,jiang2024mixtral}. Most of these are released in varying sizes with varying trade-offs between capabilities and computational efficiency.

The application of \acp{LLM} to optimisation is less prominent in recent research than other applications. Naturally fitting the \ac{NLP} origins of \acp{LLM}, they have successfully been applied to optimising prompts to \acp{LLM} chatbots~\citep{yang2023large}.
In further work, \acp{LLM} have been used to find more effective algorithms than the state of the art to solve complex problems~\citep{romeraparedes2024mathematical}.
Most similar to our work, the ability of \acp{LLM} to solve numerical optimisation has been demonstrated on the simple task of linear regression in \cite{yang2023large}.
A benchmark evaluating the performance of different \acp{LLM} on a game playing task like those typically solved by training \ac{NN} policies through \ac{RL} is presented in \cite{oulianov2024evaluate}.

In the context of particle accelerators, there exist ambitions to harness the \ac{NLP} abilities of \acp{LLM} for various purposes. In \cite{sulc2023pacuna}, the authors demonstrate how to fine-tune an open-source \ac{LLM} to be a particle accelerator domain expert using open access scientific literature as training data, augmented by another \ac{LLM} to generate question-answer pairs from research papers. The fine-tuned model, called \textit{PACuna}, is shown to be more proficient in answering questions related to particle accelerators.
In \cite{mayet2024building,mayet2024gaia}, the author demonstrates how off-the-shelf \acp{LLM} can be used as a \ac{GAIA}, employing the \textit{ReAct}~\citep{yao2023react} prompting scheme to enable the \ac{LLM} to intelligently trigger accelerator operation routines, automatically contact experts when needed, research questions in the facility's logbook, provide the correct control system addresses for actuators and sensors of the accelerator, and write weekly shift reports.

\section{Tuning Particle Accelerators Through Natural Language}\label{sec:tuning_particle_accelerators_with_llms}

For the purpose of this work, we consider a specific particle accelerator tuning task, namely the transverse beam parameter tuning in the \ac{EA} section of the \ac{ARES} linear particle accelerator~\citep{panofski2021commissioning,burkart2022the} at DESY in Hamburg, Germany.
This task has been chosen as it is a well-defined and well-understood task in the accelerator community, and has been extensively studied in the context of autonomous accelerator tuning~\citep{kaiser2022learningbased,kaiser2023learning,kaiser2023cheetah,xu2023beam}.
At the same time, the task is complex enough to be difficult to solve manually and can provide a meaningful benchmark for the capabilities of \acp{LLM} in accelerator tuning, yet simple enough such that solutions can still be easily understood and evaluated.
Solving it would provide a valuable streamlining of accelerator operations because similar transverse tuning tasks can be found at most accelerator facilities and have to be regularly performed during everyday operations.

\begin{figure}
    \centering
    \includegraphics{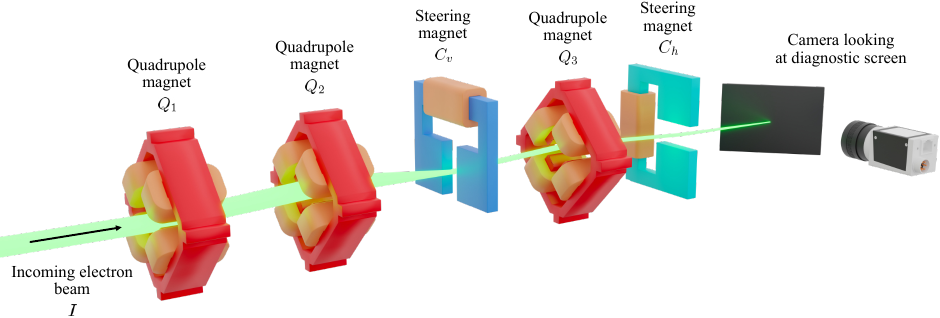}
    \caption{Schematic of the \ac{EA} section of the \ac{ARES} linear particle accelerator. Quadrupole magnets are shown in red; the vertical and horizontal dipole are shown in blue and turquoise, respectively. The electron beam is shown as a green envelop passing through the magnets and onto the screen at the end of the experimental area.}
    \label{fig:ares_ea}
\end{figure}

The \ac{EA} section is primarily made up of five magnets as shown in \cref{fig:ares_ea}. Three of these magnets are quadrupole magnets $Q_1$, $Q_2$ and $Q_3$, which are used to focus the beam, and two are dipole magnets $C_v$ and $C_h$, which are used to bend the beam, one in the vertical and one in the horizontal plane.
In this work, we control the focusing strength $k$ of the quadrupole magnets in \unit{\meter^{-2}} and the angle $\alpha$ by which the dipole magnets deflect particles in \unit{\milli\radian}.
Note that turning up the strength of a quadrupole magnet will focus the beam in the horizontal plane and defocus it in the vertical plane, while turning down the strength will have the opposite effect.
Increasing the steering angle of the vertical steering magnet will steer the beam upwards, while decreasing the angle will steer the beam downwards. The horizontal steering magnet works similarly, steering the beam to the right when the angle is increased and to the left when the angle is decreased.
What is more, quadrupole magnets also have a dipole effect on the beam, if the beam passes through the off-centre, making any tuning task involving them more complex.
The magnets are arranged in the order $\left( Q_1, Q_2, C_v, Q_3, C_h \right)$.
At the end of the \ac{EA} section, there is a diagnostic screen station. At the screen station, a screen made of a scintillating material is inserted into the beam pipe. When electrons pass through the screen, light is emitted, which is then captured by a camera and used to measure a transverse projection of the beam. Transverse beam parameters of beam position $\mu_{x,y}$ and beam size $\sigma_{x,y}$ can then be computed from the screen image by fitting a 2D Gaussian distribution.
The  goal of the tuning task is to find a set of magnet settings $\left( k_{Q_1}, k_{Q_2}, \alpha_{C_v}, k_{Q_3}, \alpha_{C_h} \right)$ that minimise the difference between the measured beam parameters $\left( \mu_x, \sigma_y, \mu_y, \sigma_y \right)$ and some target beam parameters $\left( \mu'_x, \sigma'_y, \mu'_y, \sigma'_y \right)$ set by the human operator.

\subsection{Optimisation Scheme}

\begin{figure}
    \centering
    \includegraphics[width=0.68\textwidth]{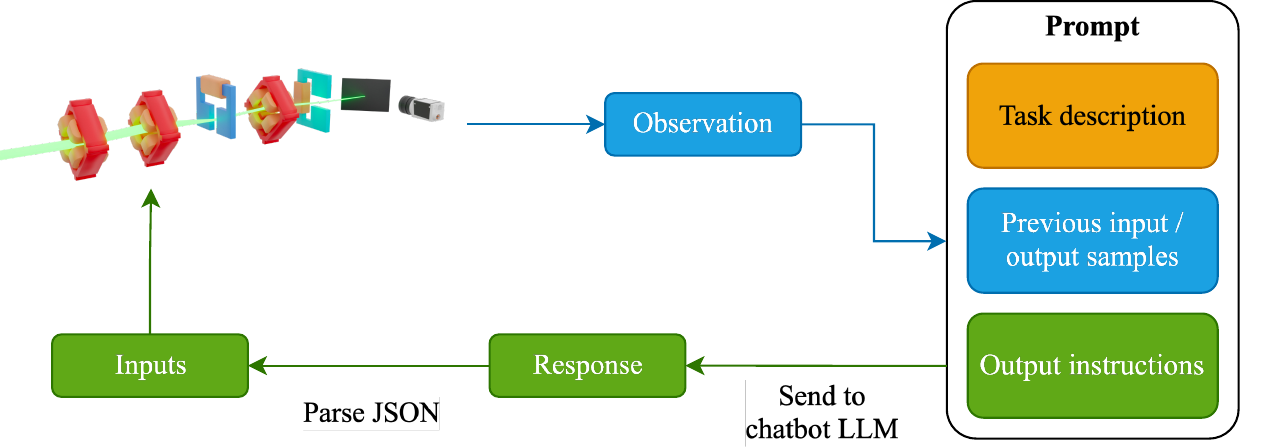}
    \caption{Flowchart of the optimisation scheme used to tune particle accelerators using \acp{LLM}. The prompt is made up for three components: Task description, list of previous input and output samples, and instructions what to output and how to format the output. The prompt is then sent to the \ac{LLM}, which generates a response. The response is parsed into values that can be input into the tuning / optimisation task. A measurement or objective value from the task is then inserted into the previous samples along with its corresponding input and the loop is repeated.}
    \label{fig:prompting_flowchart}
\end{figure}

We consider an iterative optimisation scheme for accelerator tuning, where the state of the accelerator is observed and then the tuning algorithm chooses new actuator settings based on the current and all past states from the tuning run. This process is repeated either for a fixed number of iterations or until some termination criterion is met.
For an \ac{LLM} to act as the tuning or optimisation algorithm, a prompting scheme needs to be devised. In our approach we consider the use of a chatbot \ac{LLM}, where the user can provide a question or command to the \ac{LLM} and the \ac{LLM} will respond with an answer.
Our optimisation scheme using \acp{LLM} extends on the approach for linear regression presented in \cite{yang2023large} and is shown in \cref{fig:prompting_flowchart}.
In the prompt to the \ac{LLM}, the user provides a description of the optimisation task that the \ac{LLM} should solve. This is followed by a list of input and output pairs from previous optimisation steps. After this list, the user asks for the next set of input parameters that help optimise the objective function and gives the \ac{LLM} instructions on how these should be formatted such that the user can parse the output from text to numerical values.
This prompt is then sent to the \ac{LLM}, which responds with the next set of input parameters that should be used to optimise the objective function, and potentially also an explanation of why these parameters were chosen.
The response should look similar to the one below:

\begin{promptbox}
    \`{}\`{}\`{}json

    \{

    \quad "Q1": -14.30,

    \quad "Q2": -9.70,

    \quad "CV": -2.55,

    \quad "Q3": -8.10,

    \quad "CH": -5.21

    \}

    \`{}\`{}\`{}

    I suggest decreasing Q1 slightly to bring down the horizontal beam position, while keeping the other quadrupole magnets at their previous values to maintain the vertical beam position and focusing. I also kept the steering magnet settings close to their last values for smoothness.
\end{promptbox}

The response is then parsed, and the input parameters are used to evaluate the objective function. The output of this evaluation is then added to the list along with its corresponding input parameters, and the process is repeated.

Prompt engineering is a crucial part of using \acp{LLM} and can significantly impact the performance of the model.
Because of the variability in the performance of different prompts and the difficulty of finding the best prompt, we evaluate the ability of \acp{LLM} to solve the accelerator tuning task using
four different prompts: \textit{Tuning Prompt} (see \cref{sec:tuning_prompt}), \textit{Explained Prompt} (see \cref{sec:explained_prompt}), \textit{Chain-of-Thought Prompt} (see \cref{sec:chain_of_thought_prompt}) and \textit{Optimisation Prompt} (see \cref{sec:optimisation_prompt}).
All prompts follow the general prompting scheme described above, of task description, input-output pairs, request for next input parameters and instructions on how to format the output. The prompts used in this work differ mainly in the task description and the outputs of the previous optimisation steps.

\subsubsection{Tuning Prompt}\label{sec:tuning_prompt}

The \textit{Tuning Prompt} is the most straightforward and intuitive prompt used in this work. It describes the task of tuning the transverse beam parameters in the \ac{EA} section and the goal of achieving some target beam parameters on the diagnostic screen, such that the \ac{LLM} is aware of the fact it is tuning a particle accelerator. The input-output pairs are the magnet settings and the corresponding measured beam parameters.
This prompt assumes that the \ac{LLM} has some understanding of particle accelerators and understands, for example, what a quadrupole magnet is and how it affects the beam.
Below is an example of the Tuning Prompt, where the task description is highlighted in orange, the input-output pairs in blue, and the request for the next input parameters and output instructions in green:

\begin{promptbox}
    \hlc[orange!50!white]{Human: Now you will help me optimise the horizontal and vertical position and size of an electron beam on a diagnostic screen in a particle accelerator.}

    \vspace{\baselineskip}

    \hlc[orange!50!white]{You are able to control five magnets in the beam line. The magnets are called Q1, Q2, CV, Q3 and CH.}

    \vspace{\baselineskip}

    \hlc[orange!50!white]{Q1, Q2 and Q3 are quadrupole magnets. You are controlling their k1 strength in m\^{}-2. Their range is -30.0 to 30.0 m\^{}-2.}

    \vspace{\baselineskip}

    \hlc[orange!50!white]{CV is vertical steering magnet. You control its steering angle in mrad. Its range is -6.0 to 6.0 mrad.}

    \vspace{\baselineskip}

    \hlc[orange!50!white]{CH is horizontal steering magnet. You control its steering angle in mrad. Its range is -6.0 to 6.0 mrad.}

    \vspace{\baselineskip}

    \hlc[orange!50!white]{You are optimising four beam parameters: mu\_x, sigma\_x, mu\_y, sigma\_y. The beam parameters are measured in millimetres (mm). The target beam parameters are:}

    \vspace{\baselineskip}

    \hlc[orange!50!white]{Target beam parameters:

    \`{}\`{}\`{}json

    \{

    \quad "mu\_x": 1.20,

    \quad "sigma\_x": 0.11,

    \quad "mu\_y": 1.25,

    \quad "sigma\_y": 0.06

    \}

    \`{}\`{}\`{}}

    \vspace{\baselineskip}

    \hlc[CornflowerBlue!50!white]{Below are previously measured pairs of magnet settings and the corresponding observed beam parameters.}

    \vspace{\baselineskip}

    \hlc[CornflowerBlue!50!white]{Magnet settings:

    \`{}\`{}\`{}json

    \{

    \quad "Q1": 25.12,

    \quad "Q2": 12.48,

    \quad "CV": 0.84,

    \quad "Q3": -8.25,

    \quad "CH": 3.94

    \}

    \`{}\`{}\`{}}

    \vspace{\baselineskip}

    \hlc[CornflowerBlue!50!white]{Beam parameters:

    \`{}\`{}\`{}json

    \{

    \quad "mu\_x": -1038.63,

    \quad "sigma\_x": 1893.75,

    \quad "mu\_y": -2353.77,

    \quad "sigma\_y": 2226.94

    \}

    \`{}\`{}\`{}}

    \vspace{\baselineskip}

    \hlc[YellowGreen!80!white]{Give me new magnet settings that are different from all pairs above. The magnet settings you should propose should lead to beam parameters closer the target or, if you do not have enough information yet, maximise information gain for finding new beam parameters. Do not set any magnet setting to zero. Smooth changes relative to the last magnet settings are preferred.}

    \vspace{\baselineskip}

    \hlc[YellowGreen!80!white]{The output should be a markdown code snippet formatted in the following schema, including the leading and trailing "\`{}\`{}\`{}json" and "\`{}\`{}\`{}":}

    \vspace{\baselineskip}

    \hlc[YellowGreen!80!white]{\`{}\`{}\`{}

    \{

    \quad "Q1": float  // k1 strength of the first quadrupole magnet

    \quad "Q2": float  // k1 strength of the second quadrupole magnet

    \quad "CV": float  // Deflection angle of the vertical steering magnet

    \quad "Q3": float  // k1 strength of the third quadrupole magnet

    \quad "CH": float  // Deflection angle of the horizontal steering magnet

    \}

    \`{}\`{}\`{}}

    \vspace{\baselineskip}

    \hlc[YellowGreen!80!white]{Do not add comments to the output JSON.}
\end{promptbox}

Note that the choice was made to provide previously observed magnet settings and beam parameters formatted as a markdown JSON snippet.
We found that if these are provided as a simple textual list of property names and their values, the \acp{LLM} would often output the next magnet settings in the same format instead of the requested JSON format. By providing the examples in the same format as we desire for the output, the parsing reliability of the \ac{LLM} is increased significantly.

\subsubsection{Explained Prompt}\label{sec:explained_prompt}

The \textit{Explained Prompt} is mostly the same as the Tuning Prompt, but includes additional explanations of how each of the magnets affects the beam.
This is done because accelerator physics is a complex and niche field, which is unlikely to have been widely covered in the training data of a general-purpose \acp{LLM}.
The explanations are generally kept on a high level, similar to how one might explain the task to a new accelerator operator in order to give them an intuition of how the magnets affect the beam on the diagnostic screen.
Below is an example of the Explained Prompt with the explanations added over the Tuning Prompt highlighted in violet:

\begin{promptbox}
    Human: Now you will help me optimise the horizontal and vertical position and size of an electron beam on a diagnostic screen in a particle accelerator.

    \vspace{\baselineskip}

    You are able to control five magnets in the beam line. The magnets are called Q1, Q2, CV, Q3 and CH.

    \vspace{\baselineskip}

    Q1, Q2 and Q3 are quadrupole magnets. \hlc[violet!33!white]{When their k1 strength is increased, the beam becomes more focused in the horizontal plane and more defocused in the vertical plane. When their k1 strength is decreased, the beam becomes more focused in the vertical plane and more defocused in the horizontal plane. When their k1 strength is zero, the beam is not focused in either plane. Quadrupole magnets might also steer the beam in the horizontal or vertical plane depending on their k0 strength, when the beam does not travel through the centre of the magnet. The range of the k1 strength is -30.0 to 30.0 m\^{}-2.}

    \vspace{\baselineskip}

    CV is vertical steering magnet. \hlc[violet!33!white]{When its deflection angle is increased, the beam is steered upwards. When its deflection angle is decreased, the beam is steered downwards. The range of the deflection angle is -6.0 to 6.0 mrad.}

    \vspace{\baselineskip}

    CH is horizontal steering magnet. \hlc[violet!33!white]{When its deflection angle is increased, the beam is steered to the right. When its deflection angle is decreased, the beam is steered to the left. The range of the deflection angle is -6.0 to 6.0 mrad.}

    \vspace{\baselineskip}

    You are optimising four beam parameters: mu\_x, sigma\_x, mu\_y, sigma\_y. The beam parameters are measured in millimetres (mm). The target beam parameters are:

    \vspace{\baselineskip}

    Target beam parameters:

    \`{}\`{}\`{}json

    \{

    \quad "mu\_x": 1.20,

    \quad "sigma\_x": 0.11,

    \quad "mu\_y": 1.25,

    \quad "sigma\_y": 0.06

    \}

    \`{}\`{}\`{}

    \vspace{\baselineskip}

    Below are previously measured pairs of magnet settings and the corresponding observed beam parameters.

    \vspace{\baselineskip}

    Magnet settings:

    \`{}\`{}\`{}json

    \{

    \quad "Q1": 25.12,

    \quad "Q2": 12.48,

    \quad "CV": 0.84,

    \quad "Q3": -8.25,

    \quad "CH": 3.94

    \}

    \`{}\`{}\`{}

    Beam parameters:

    \`{}\`{}\`{}json

    \{

    \quad "mu\_x": -1038.63,

    \quad "sigma\_x": 1893.75,

    \quad "mu\_y": -2353.77,

    \quad "sigma\_y": 2226.94

    \}

    \`{}\`{}\`{}

    \vspace{\baselineskip}

    Give me new magnet settings that are different from all pairs above. The magnet settings you should propose should lead to beam parameters closer the target or, if you do not have enough information yet, maximise information gain for finding new beam parameters. Do not set any magnet setting to zero. Smooth changes relative to the last magnet settings are preferred.

    \vspace{\baselineskip}

    The output should be a markdown code snippet formatted in the following schema, including the leading and trailing "\`{}\`{}\`{}json" and "\`{}\`{}\`{}":

    \vspace{\baselineskip}

    \`{}\`{}\`{}json

    \{

    \quad "Q1": float  // k1 strength of the first quadrupole magnet

    \quad "Q2": float  // k1 strength of the second quadrupole magnet

    \quad "CV": float  // Deflection angle of the vertical steering magnet

    \quad "Q3": float  // k1 strength of the third quadrupole magnet

    \quad "CH": float  // Deflection angle of the horizontal steering magnet

    \}

    \`{}\`{}\`{}

    \vspace{\baselineskip}

    Do not add comments to the output JSON.
\end{promptbox}

\subsubsection{Chain-of-Thought Prompt}\label{sec:chain_of_thought_prompt}

\Acf{CoT} prompting~\citep{wei2023chainofthought} is a technique where the user asks the \ac{LLM} to explain its reasoning before it gives its answer.
This was found to generally improve the quality of the answers given by \acp{LLM}, especially in the case of logical reasoning tasks.
Note that it is important to have the explanation before the answer, as otherwise the model will phrase the explanation in support of the already given and potentially incorrect answer, thereby negating the benefit of chain-of-thought prompting.
In the \textit{Chain-of-Thought Prompt}, we add a request to the prompt whereby the users asks the \ac{LLM} to explain its reasoning before it gives the next set of input parameters.
Otherwise, the Chain-of-Thought Prompt is the same as the Explained Prompt.
Below is an example of the Chain-of-Thought Prompt, where the request for chain-of-thought reasoning is highlighted in violet:

\begin{promptbox}
    Human: Now you will help me optimise the horizontal and vertical position and size of an electron beam on a diagnostic screen in a particle accelerator.

    \vspace{\baselineskip}

    You are able to control five magnets in the beam line. The magnets are called Q1, Q2, CV, Q3 and CH.

    \vspace{\baselineskip}

    Q1, Q2 and Q3 are quadrupole magnets. When their k1 strength is increased, the beam becomes more focused in the horizontal plane and more defocused in the vertical plane. When their k1 strength is decreased, the beam becomes more focused in the vertical plane and more defocused in the horizontal plane. When their k1 strength is zero, the beam is not focused in either plane. Quadrupole magnets might also steer the beam in the horizontal or vertical plane depending on their k0 strength, when the beam does not travel through the centre of the magnet. The range of the k1 strength is -30.0 to 30.0 m\^{}-2.

    \vspace{\baselineskip}

    CV is vertical steering magnet. When its deflection angle is increased, the beam is steered upwards. When its deflection angle is decreased, the beam is steered downwards. The range of the deflection angle is -6.0 to 6.0 mrad.

    \vspace{\baselineskip}

    CH is horizontal steering magnet. When its deflection angle is increased, the beam is steered to the right. When its deflection angle is decreased, the beam is steered to the left. The range of the deflection angle is -6.0 to 6.0 mrad.

    \vspace{\baselineskip}

    You are optimising four beam parameters: mu\_x, sigma\_x, mu\_y, sigma\_y. The beam parameters are measured in millimetres (mm). The target beam parameters are:

    \vspace{\baselineskip}

    Target beam parameters:

    \`{}\`{}\`{}json

    \{

    \quad "mu\_x": 1.20,

    \quad "sigma\_x": 0.11,

    \quad "mu\_y": 1.25,

    \quad "sigma\_y": 0.06

    \}

    \`{}\`{}\`{}

    \vspace{\baselineskip}

    Below are previously measured pairs of magnet settings and the corresponding observed beam parameters.

    \vspace{\baselineskip}

    Magnet settings:

    \`{}\`{}\`{}json

    \{

    \quad "Q1": 25.12,

    \quad "Q2": 12.48,

    \quad "CV": 0.84,

    \quad "Q3": -8.25,

    \quad "CH": 3.94

    \}

    \`{}\`{}\`{}

    Beam parameters:

    \`{}\`{}\`{}json

    \{

    \quad "mu\_x": -1038.63,

    \quad "sigma\_x": 1893.75,

    \quad "mu\_y": -2353.77,

    \quad "sigma\_y": 2226.94

    \}

    \`{}\`{}\`{}

    \vspace{\baselineskip}

    Give me new magnet settings that are different from all pairs above. The magnet settings you should propose should lead to beam parameters closer the target or, if you do not have enough information yet, maximise information gain for finding new beam parameters. Do not set any magnet setting to zero. Smooth changes relative to the last magnet settings are preferred.

    \vspace{\baselineskip}

    \hlc[violet!33!white]{First, reason about how and why you would change the magnet settings in a certain direction. Then give me the proposed magnet settings afterwards.}

    \vspace{\baselineskip}

    The output should be a markdown code snippet formatted in the following schema, including the leading and trailing "\`{}\`{}\`{}json" and "\`{}\`{}\`{}":

    \vspace{\baselineskip}

    \`{}\`{}\`{}json

    \{

    \quad "Q1": float  // k1 strength of the first quadrupole magnet

    \quad "Q2": float  // k1 strength of the second quadrupole magnet

    \quad "CV": float  // Deflection angle of the vertical steering magnet

    \quad "Q3": float  // k1 strength of the third quadrupole magnet

    \quad "CH": float  // Deflection angle of the horizontal steering magnet

    \}

    \`{}\`{}\`{}

    \vspace{\baselineskip}

    Do not add comments to the output JSON.
\end{promptbox}

\subsubsection{Optimisation Prompt}\label{sec:optimisation_prompt}

The \textit{Optimisation Prompt} phrases the task as a numerical optimisation problem instead of a particle accelerator tuning task.
This means that the model is completely unaware that it is tuning a particle accelerator.
Numerical optimisation tasks are more generic than particle accelerator tuning tasks and therefore expected to be more present in the training data used to train \acp{LLM}, meaning that models are likely to be more familiar with them. However,
this also means that the model is given no information about the topology of the objective function, which makes the optimisation problem harder to solve.
The objective function is therefore a black box to the model.
The input-output pairs are the magnet settings and a corresponding single scalar objective value computed from the beam parameters as

% Mean of the differences between the absolute values of all four beam parameters
\begin{equation}
    \mathrm{objective} = \left| \mu_x - \mu_x' \right| + \left| \mu_y - \mu_y' \right| + \left| \sigma_x - \sigma_x' \right| + \left| \sigma_y - \sigma_y' \right| \text{.}
\end{equation}

Below is an example of the Optimisation Prompt, where the task description is highlighted in orange, the input-output pairs in blue, and the request for the next input parameters and output instructions in green:

\begin{promptbox}
    \hlc[orange!50!white]{Human: Now you will help me minimise a function with five input variables Q1, Q2, CV, Q3 and CH. I have some (Q1, Q2, CV, Q3, CH) pairs and the corresponding function values at those points. The samples are arranged in descending order based on their function values, where lower values are better.}

    \vspace{\baselineskip}

    \hlc[CornflowerBlue!50!white]{Inputs:

    \`{}\`{}\`{}json

    \{

    \quad "Q1": -13.50,

    \quad "Q2": -9.00,

    \quad "CV": -3.00,

    \quad "Q3": -9.00,

    \quad "CH": -6.00

    \}

    \`{}\`{}\`{}

    Objective value = 2.37}

    \vspace{\baselineskip}

    \hlc[CornflowerBlue!50!white]{Inputs:

    \`{}\`{}\`{}json

    \{

    \quad "Q1": -13.25,

    \quad "Q2": -8.85,

    \quad "CV": -2.80,

    \quad "Q3": -8.90,

    \quad "CH": -5.70

    \}

    \`{}\`{}\`{}

    Objective value = 2.28}

    \vspace{\baselineskip}

    \hlc[YellowGreen!80!white]{Give me a new sample (Q1, Q2, CV, Q3, CH) that is different from all pairs above, and has a function value lower than any of the above.}

    \vspace{\baselineskip}

    \hlc[YellowGreen!80!white]{The output should be a markdown code snippet formatted in the following schema, including the leading and trailing "\`{}\`{}\`{}json" and "\`{}\`{}\`{}":}

    \vspace{\baselineskip}

    \hlc[YellowGreen!80!white]{\`{}\`{}\`{}json

    \{

    \quad "Q1": float  // First input

    \quad "Q2": float  // Second input

    \quad "CV": float  // Third input

    \quad "Q3": float  // Fourth input

    \quad "CH": float  // Fifth input

    \}

    \`{}\`{}\`{}}
\end{promptbox}

\section{Evaluation}\label{sec:evaluation}

In order to evaluate whether \acp{LLM} using the prompting scheme described in \cref{sec:tuning_particle_accelerators_with_llms} are capable of solving particle accelerator tuning tasks, we compare the performance of multiple state-of-the-art \acp{LLM} against each other and against other state-of-the-art accelerator tuning solutions using \ac{RLO} and \ac{BO}.
In addition, we consider some further baselines for our comparison, specifically \ac{ES}, random search and doing nothing.
The evaluation setup is introduced in \cref{sec:evaluation_method}, followed by the evaluation results in \cref{sec:evaluation_results}.

\subsection{Method}\label{sec:evaluation_method}

We evaluate each of the models and prompts on three different instances of the \ac{EA} transverse beam parameter tuning task described in \cref{sec:tuning_particle_accelerators_with_llms}.
We call these instances \textit{trials}.
Trials differ in the target beam parameters set by the human operator, the transverse misalignments of the quadrupole magnets and the diagnostic screen, the properties of the beam entering the \ac{EA} section from upstream, and the initial magnet settings before the respective tuning algorithm has taken any action.
For each trial, we run each model and prompt three times with different random seeds to account for the stochasticity of the \acp{LLM} and some of the baseline algorithms.

Performance is evaluated in terms of the \ac{MAE} between the measured beam parameters and the target beam parameters after \num{50} iterations. This tests the ability of the models to find a good set of magnet settings.
We further consider the normalised \ac{MAE} improvement from the initial magnet settings to the final magnet settings found by the model, which tests the ability of the models to improve the beam parameters from the initial settings.
Normalisation by dividing the \ac{MAE} improvement by the \ac{MAE} with the initial magnet settings makes this metric less sensitive to the inherent variability and difficulty of different trials.
Finally, we consider the normalised \ac{MAE} over all interactions, which tests the ability of the models to find a good set of magnet settings quickly.
Here, too, the impact of trial-to-trial variations is reduced by dividing by the accumulated \ac{MAE} of keeping the magnet settings the same as the initial settings for \num{50} iterations.
For all \acp{LLM}, we further consider the number of consecutive steps for which they are able to generate a parsable JSON output, which tests the tests the models' reliability in generating a valid output.
\Acp{LLM} are given a second chance in each sample, if they fail to generate a parsable JSON output on the first attempt.

The main goal of this work is not to determine whether \acp{LLM} are capable of outperforming the current state of the art in accelerator tuning algorithms. In fact, we expect that the current state of the art in accelerator tuning algorithms, such as \ac{RLO} and \ac{BO}, clearly outperform \acp{LLM}. Instead, we hope to determine whether \acp{LLM} are capable of solving accelerator tuning (and by extension other complex optimisation tasks) at all, and to what extent they can do so.
We therefore also introduce three measures of ``success'', where we consider a tuning run successful, if the final beam difference is at least is \qty{40}{\micro\meter} improved over the initial beam difference before any tuning has taken place,
with \qty{40}{\micro\meter} being twice the real-world measurement accuracy for beam parameters on the diagnostic screen. This means that runs are only considered successful, if a clearly measurable improvement of the beam parameters has been achieved.
A tuning algorithm is considered ``outright successful'', if it is able to achieve the success criteria in all \num{9} evaluation runs. We consider a tuning algorithm as ``partially successful'' if it is able to achieve the success criterion in at least \num{6} of the \num{9} evaluation runs. Partial success suggests that, while not perfectly reliable, successful runs probably not coincidental.
We further know that some trials can be harder to solve than others.
As a third and weakest success criterion, we therefore consider a tuning algorithm as ``single trial successful'' if it is able to achieve the success criterion in on all three runs of a single trial, suggesting that, while some trials may have been too difficult to solve, the model was able to reliably solve this one trial.

For this work we evaluate a total of \num{14} different \acp{LLM}.
These are \textit{Gemma 2B} and \textit{Gemma 7B}~\citep{gemmateam2024gemma} (version 1.0);
\textit{GPT 3.5 Turbo} (checkpoint 0125)~\citep{openai2023gpt35turbo}, \textit{GPT 4}~\citep{openai2023gpt4} (checkpoint 0613) and \textit{GPT 4 Turbo} (preview checkpoint 0125)~\citep{openai2023new};
\textit{Llama 2 7B}, \textit{Llama 2 13B} and \textit{Llama 2 70B}~\citep{touvron2023llama}, as well as the fine-tuned variants of Llama 2: \textit{Orca 2 7B} and \textit{Orca 2 13B}~\citep{mukherjee2023orca,mitra2023orca}, and \textit{Vicuna 7B 16K}~\citep{zheng2023judging};
\textit{Mistral 7B} (version 0.2)~\citep{jiang2023mistral} and \textit{Mixtral 8x7B}~\citep{jiang2024mixtral};
and \textit{Starling-LM 7B} (beta)~\citep{zhu2023starling}.
The Explained Prompt and Optimisation Prompt are evaluated with all models, while the Tuning Prompt and Chain-of-Thought Prompt are evaluated only with Gemma 2B, GPT 4 Turbo and Mixtral 8x7B.

Prompt generation and response parsing are implemented using the \textit{LangChain}~\citep{chase2022langchain} Python package, which provides a straightforward set of tools for constructing prompts, calling \acp{LLM} and parsing their responses.
The open-weights \acp{LLM} used in this work are run using \textit{Ollama}~\citep{ollamateam2024ollama}, while the OpenAI models are run through the OpenAI API.
All models are run using their default temperature value, with $T = \num{0.7}$ for the OpenAI models and $T = \num{0.8}$ for all other models.
Orca 2 7B, Orca 2 13B and Vicuna 7B 16K are run with their default system prompts as listed in \cref{sec:system_prompts}. All other models are run without any system prompts as per their default configuration.
A \textit{Gymnasium}~\citep{farama2022gymnasium} environment of the \ac{EA} transverse beam parameter tuning task using the \textit{Cheetah}~\citep{stein2022accelerating,kaiser2024cheetah} beam dynamics simulator is used to evaluate the \acp{LLM} and baselines.
The baselines \ac{BO}, \ac{ES} and random search are implemented following \cite{kaiser2023learning}. The \ac{RLO} and do nothing baselines are implemented according to \cite{kaiser2022learningbased}, using the trained policy model from that work.

\subsection{Results}\label{sec:evaluation_results}

\begin{sidewaystable}
    \caption{Evaluation Results}
    \centering
    \begin{tabular}{lccccccccc}
        \toprule
                         & \multicolumn{2}{c}{Final beam difference (\unit{\micro\meter})} & \multicolumn{2}{c}{Normalised beam improvement (\unit{\percent})} & \multicolumn{2}{c}{Normalised integrated MAE (\unit{\percent})} & \multicolumn{2}{c}{Number of successful steps}                                                                           \\
        \cmidrule(r){2-3} \cmidrule(r){4-5} \cmidrule(r){6-7} \cmidrule(r){8-9}
                         & Explained                                                       & Optimisation                                                      & Explained                                                       & Optimisation                                   & Explained        & Optimisation     & Explained       & Optimisation    \\
        \midrule
        Gemma 2B         & 1665 ± 634                                                      & 3180 ± 5187                                                       & 11 ± 71                                                         & 34 ± 171                                       & 115 ± 51         & 137 ± 88         & 23 ± 19         & 39 ± 14         \\
        Gemma 7B         & 1651 ± 764                                                      & 8105 ± 12933                                                      & -16 ± 11                                                        & 284 ± 428                                      & 85 ± 10          & 247 ± 142        & 9 ± 0           & 29 ± 11         \\
        GPT 3.5 Turbo    & 11593 ± 14850                                                   & 1197 ± 771                                                        & 397 ± 618                                                       & -36 ± 27                                       & 292 ± 245        & 78 ± 20          & \textbf{50} ± 0 & \textbf{50} ± 0 \\
        GPT 4            & 1849 ± 1445                                                     & 1213 ± 860                                                        & 11 ± 73                                                         & -40 ± 25                                       & 98 ± 60          & 73 ± 20          & \textbf{50} ± 0 & \textbf{50} ± 0 \\
        GPT 4 Turbo      & 2184 ± 1879                                                     & \textbf{962} ± 740                                                & 20 ± 89                                                         & \textbf{-50} ± 28                              & 107 ± 76         & \textbf{67} ± 21 & \textbf{50} ± 0 & \textbf{50} ± 0 \\
        Llama 2 7B       & 1432 ± 798                                                      & 2085 ± 779                                                        & -12 ± 55                                                        & 15 ± 27                                        & 94 ± 15          & 106 ± 15         & 8 ± 6           & 3 ± 4           \\
        Llama 2 13B      & 1936 ± 772                                                      & 1507 ± 821                                                        & 5 ± 26                                                          & -22 ± 23                                       & 101 ± 10         & 85 ± 21          & 0 ± 1           & 13 ± 20         \\
        Llama 2 70B      & 1947 ± 964                                                      & 1539 ± 942                                                        & 10 ± 42                                                         & -21 ± 27                                       & 107 ± 37         & 92 ± 16          & \textbf{50} ± 0 & \textbf{50} ± 0 \\
        Orca 2 7B        & 2149 ± 1222                                                     & 1377 ± 855                                                        & 17 ± 42                                                         & -23 ± 37                                       & 122 ± 65         & 93 ± 17          & 4 ± 3           & 4 ± 7           \\
        Orca 2 13B       & 1634 ± 875                                                      & 3232 ± 3684                                                       & -13 ± 24                                                        & 77 ± 170                                       & 88 ± 18          & 142 ± 92         & 1 ± 2           & 3 ± 2           \\
        Vicuna 7B 16K    & 4756 ± 5332                                                     & 4331 ± 3829                                                       & 184 ± 320                                                       & 320 ± 580                                      & 189 ± 137        & 234 ± 236        & 34 ± 7          & 48 ± 7          \\
        Mistral 7B       & 2551 ± 1233                                                     & 19653 ± 23427                                                     & 48 ± 57                                                         & 803 ± 869                                      & 121 ± 40         & 1574 ± 1513      & \textbf{50} ± 0 & 30 ± 22         \\
        Mixtral 8x7B     & 1606 ± 1158                                                     & 1901 ± 1192                                                       & \textbf{-24} ± 27                                               & -14 ± 31                                       & \textbf{76} ± 26 & 101 ± 26         & \textbf{50} ± 0 & 45 ± 14         \\
        Starling LM 7B   & \textbf{1401} ± 449                                             & 7659 ± 7249                                                       & 1 ± 69                                                          & 363 ± 521                                      & 98 ± 60          & 324 ± 252        & 36 ± 15         & 19 ± 16         \\
        \midrule
                         & Tuning                                                          & \Ac{CoT}                                                          & Tuning                                                          & \Ac{CoT}                                       & Tuning           & \Ac{CoT}         & Tuning          & \Ac{CoT}        \\
        \midrule
        Gemma 2B         & 1452 ± 525                                                      & \textbf{955} ± 702                                                & -14 ± 46                                                        & \textbf{-40} ± 49                              & 97 ± 43          & 87 ± 60          & 10 ± 1          & \textbf{50} ± 0 \\
        GPT 4 Turbo      & 2647 ± 1827                                                     & 1337 ± 813                                                        & 45 ± 81                                                         & -23 ± 45                                       & 119 ± 64         & \textbf{70} ± 25 & \textbf{50} ± 0 & \textbf{50} ± 0 \\
        Mixtral 8x7B     & \textbf{1321} ± 771                                             & 1775 ± 926                                                        & \textbf{-29} ± 23                                               & -8 ± 17                                        & \textbf{71} ± 22 & 95 ± 19          & \textbf{50} ± 0 & \textbf{50} ± 0 \\
        \midrule
                         & \multicolumn{8}{c}{Baselines}                                                                                                                                                                                                                                                                                                    \\
        \midrule
        \Ac{RLO}         & \multicolumn{2}{c}{\textbf{16} ± 17}                            & \multicolumn{2}{c}{\textbf{-99} ± 1}                              & \multicolumn{2}{c}{\textbf{3} ± 1}                              & \multicolumn{2}{c}{-}                                                                                                    \\
        \Ac{BO}          & \multicolumn{2}{c}{100 ± 26}                                    & \multicolumn{2}{c}{-93 ± 6}                                       & \multicolumn{2}{c}{31 ± 23}                                     & \multicolumn{2}{c}{-}                                                                                                    \\
        Extremum seeking & \multicolumn{2}{c}{457 ± 267}                                   & \multicolumn{2}{c}{-71 ± 19}                                      & \multicolumn{2}{c}{35 ± 17}                                     & \multicolumn{2}{c}{-}                                                                                                    \\
        Random search    & \multicolumn{2}{c}{7677 ± 3830}                                 & \multicolumn{2}{c}{487 ± 588}                                     & \multicolumn{2}{c}{647 ± 476}                                   & \multicolumn{2}{c}{-}                                                                                                    \\
        Do nothing       & \multicolumn{2}{c}{1967 ± 903}                                  & \multicolumn{2}{c}{0 ± 0}                                         & \multicolumn{2}{c}{100 ± 0}                                     & \multicolumn{2}{c}{-}                                                                                                    \\
        \bottomrule
    \end{tabular}
    \label{tab:results}
\end{sidewaystable}

\begin{figure}
    \centering
    \includegraphics{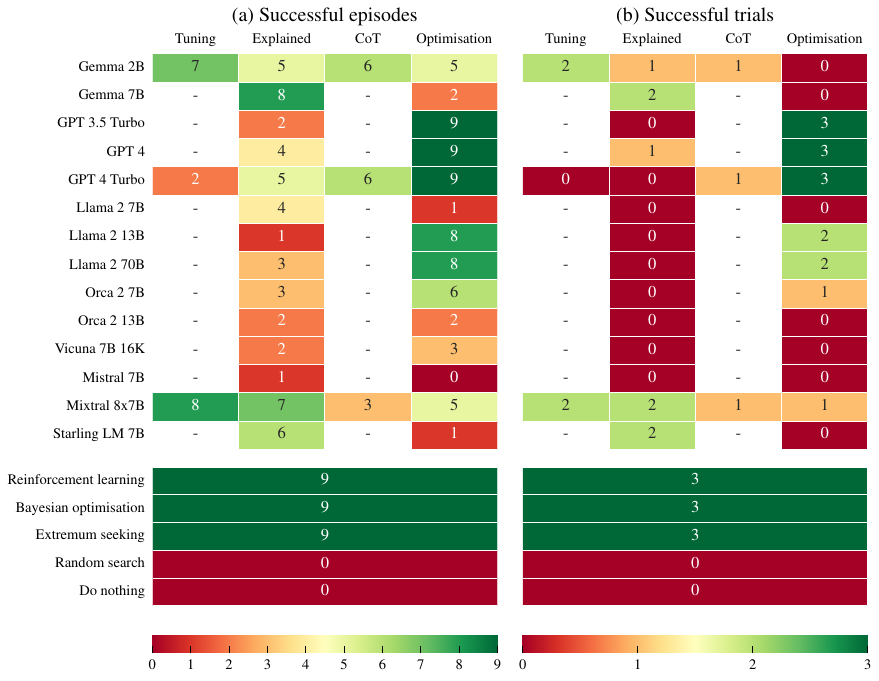}
    \caption{Number of successful runs for each model and prompt (a) and the number wholly successful trials, i.e. trials where all three runs were successful (b). We define as success an improvement of at least \qty{40}{\micro\meter} on the beam differences when compared to the initial magnet settings.}
    \label{fig:success_heatmaps}
\end{figure}

\begin{figure}
    \centering
    \includegraphics{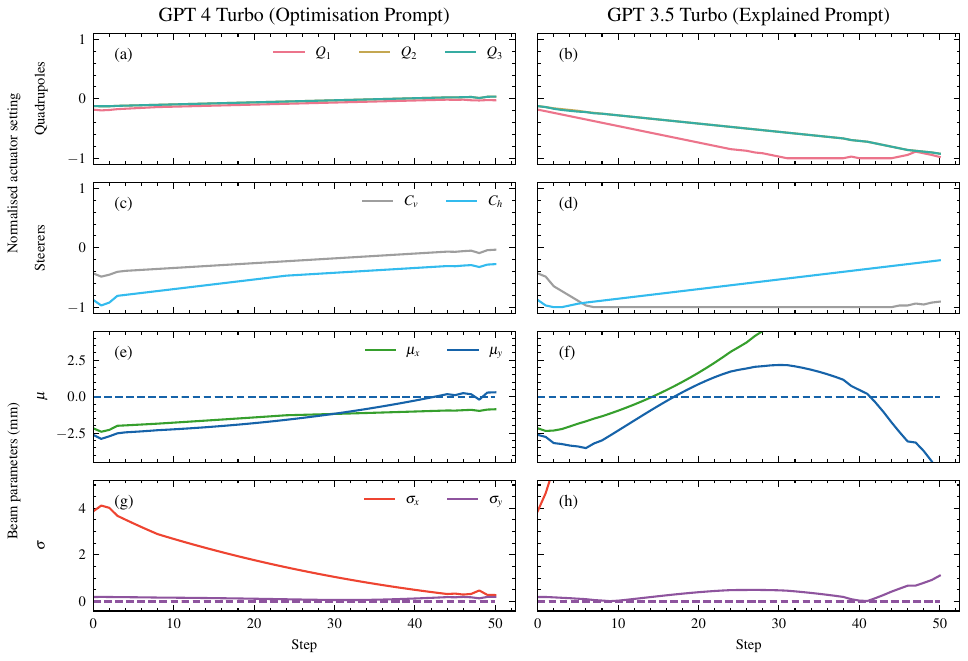}
    \caption{Magnet setting and beam parameter traces for a good and a bad tuning run by \acp{LLM}. Both runs used the same trial, where the target beam parameters are $\mu_x = \mu_y = \sigma_x = \sigma_y = \qty{0}{\milli\meter}$.}
    \label{fig:example_episodes_combined}
\end{figure}

The results of the evaluation in terms of the three previously defined metrics are shown in \cref{tab:results}. The number of successful runs and wholly successful trials for each model and prompt are shown in \cref{fig:success_heatmaps}.
Two example tuning runs by a well-performing and a poorly-performing model and prompt combination are shown in \cref{fig:example_episodes_combined}.

We find that the state-of-the-art tuning algorithms \ac{RLO} and \ac{BO}, as well as \ac{ES}, all achieve the strictest success criterion of outright success in all \num{9} evaluation runs.
Of the \ac{LLM} prompt combinations evaluated, GPT 3.5 Turbo, GPT 4 and GPT 4 Turbo in combination with the Optimisation Prompt also achieve outright success in all \num{9} evaluation runs, with GPT 4 Turbo with the Optimisation Prompt also being the best-performing \ac{LLM} prompt combination in all evaluated metrics.
In addition, a further \num{10} \ac{LLM} prompt combinations achieve partial success, with Llama 2 13B, Llama 2 70B and Orca 2 7B doing so with the Optimisation Prompt; Gemma 7B, Mixtral 8x7B and Starling LM 7B achieving partial success with the Explained Prompt; Gemma 2B and Mixtral 8x7B achieving partial success with the Tuning Prompt; and Gemma 2B and GPT 4 Turbo achieving partial success with the Chain-of-Thought Prompt.
Overall, Mixtral 8x7B is the best performing model with the Explained Prompt, but is outperformed by Starling LM 7B on the Final Beam Difference metric. With the Tuning Prompt, Mixtral 8x7B performs best of the three evaluated models, while Gemma 2B is the best-performing model with the Chain-of-Thought Prompt.
All models that achieve partial success also achieve single trial success in at least one trial, demonstrating that they are able to solve some trials reliably.
A further \num{6} \ac{LLM} prompt combinations achieve single trial success: Gemma 2B and GPT 4 with the Explained Prompt, Mixtral 8x7B with the Optimisation Prompt, and Mixtral 8x7B with the Chain-of-Thought Prompt.
In total, of the \num{34} \ac{LLM} prompt combinations tried, \num{18} achieve at least one success criterion. Of \num{14} \acp{LLM} evaluated, \num{10} achieve at least one success criterion with at least one prompt.
This demonstrates that \acp{LLM} can be used to solve accelerator tuning tasks.

However, these results also show that \acp{LLM} are not yet competitive with the state-of-the-art accelerator tuning algorithms. The best-performing \ac{LLM} prompt combination, GPT 4 Turbo with the Optimisation Prompt, achieves an average normalised beam improvement of \qty{-50}{\percent}. This is a good result, but it is also a significantly worse result than the \qty{-99}{\percent} and \qty{-93}{\percent} achieved by \ac{RLO} and \ac{BO}, respectively.
A similar trend can be observed in how fast algorithms are able to find a good set of magnet settings.
GPT 4 Turbo with the Optimisation Prompt achieves an average normalised integrated \ac{MAE} of \qty{67}{\percent}, which is an order of magnitude worse than the \qty{3}{\percent} achieved by \ac{RLO}.
However, it is only about two times worse than \ac{BO} and \ac{ES}.

What is more, the results show that the performance of \acp{LLM} is highly dependent on the specific model and prompt used. While \num{18} of the \num{34} \ac{LLM} prompt combinations tried achieve at least one success criterion, the remaining \num{16} do not achieve any.
Similarly, \num{4} of the evaluated \acp{LLM} do not achieve any success criterion with any of the prompts they were tested on.
We observe that in general, the Optimisation prompt performs best in our evaluations. Outright success was only achieved with the Optimisation Prompt, and at least one success criterion was achieved by \num{7} \acp{LLM} when using the Optimisation Prompt, while only \num{5} \acp{LLM} achieve at least one success criterion with the Explained Prompt.
The best-performing \ac{LLM} prompt combination, GPT 4 Turbo with the Optimisation Prompt, also uses the Optimisation prompt.
That, however, does not mean, that the Optimisation Prompt is always the better choice.
Some models perform better with one of the other prompts.
Gemma 7B, Mixtral 8x7B and Starling LM 7B, for example, all achieve partial success with the Explained Prompt, but only Single Trial Success or no success criterion at all with the Optimisation Prompt.
Similarly, Gemma 2B and Mixtral 8x7B achieve their best results with the Tuning Prompt.
We conclude that the choice of prompt must be made on a model-by-model basis.

It is also worth noting that adding explanations to the prompts about how the magnets work, or adding a chain-of-thought to the prompts, does not always lead to the expected improvements.
Of the three models evaluated with all four prompts, only GPT 4 Turbo improves with the addition of explanations. However, this is with GPT 4 Turbo generally performing badly on any of the three variants of the Tuning Prompt, generally performing better with the Optimisation Prompt.
Gemma 2B and Mixtral 8x7B, on the other hand, perform worse when the explanations are added.
A possible explanation for this observation is that, rather than helping the model understand the tuning task, the length of the explanations makes it harder for the \ac{LLM} to retrieve relevant information, such as specific past samples or the target beam parameters, from the prompt.
This problem is known as \textit{Needle in a Haystack} and a general challenge with \acp{LLM}.
Chain-of-thought prompting seems to improve performance over the Explained Prompt with Gemma 2B and GPT 4 Turbo, but has an adverse effect on the performance of Mixtral 8x7B.
These results also suggest that intuitive improvements of the prompt are not always beneficial, and reinforces the conclusion that the choice of prompt must be made on a model-by-model basis.

In designing the presented \ac{LLM} tuning solution, we found that aside from getting \acp{LLM} to successfully tune the particle accelerator, another difficulty is to get them to output the magnet settings in a parsable JSON format.
This is why \acp{LLM} are given a second chance in each sample, if the parsing of their response fails on the first attempt.
Nevertheless, some models fail on the second attempt as well, at which point we consider the tuning run terminated.
We can therefore take number of performed iterations (excluding second attempts) as an indicator of a model's ability to produce a valid JSON output when provided with one of our prompts.
Note that excluding second attempts, this is the number of interactions with the accelerator, not the number of times the \ac{LLM} was prompted.
The observed number of iterations for the \num{9} evaluation runs of each model and prompt are shown in \cref{tab:results}. We observe many models, often those achieving good tuning results, have a high number of successful steps, with models like those by OpenAI and Llama 70B always achieving the maximum of \num{50} successful steps, regardless of the prompt used.
Other models, such as both Orca 2 and the smallest variant of Llama 2, consistently struggle to produce a valid JSON output, with the number of successful steps being very low for either prompt.
While in most cases, it appears that the ability to generate valid JSON responses depends mostly on the \ac{LLM} used, we also observe that the choice of prompt can have an impact in a few cases, with the difference being especially pronounced for the Gemma models, which achieve a higher number of successful steps with the Optimisation Prompt than with the Explained Prompt. It does not appear as though one prompt is generally better than the other in terms of the number of successful steps.
Furthermore, the nature of different invalid responses varies greatly.
In some cases, the mistakes are so minor that human experts might fail to spot them, for example when a trailing comma is added to the last JSON property.
This is not allowed in JSON syntax and causes the parser to fail. Another failure case is related to chain-of-thought.
For example, Orca 2 -- a model specifically trained to respond with chain-of-though -- often outputs an explanation of a strategy to solve the optimisation problem rather than the next magnet settings requested in the prompt.
Last, but certainly not least, some models fail to output a coherent response altogether, with responses being nonsensical, for example starting the response with an invalid continuation of a JSON object and then continuing with multiple valid JSON objects even though only a single one was requested.
In this case, both the invalid JSON object and the ambiguity about which JSON object should be parsed, make the response invalid.
Examples of these three described failure modes are given in \cref{sec:failed_responses}.

\begin{figure}
    \centering
    \includegraphics{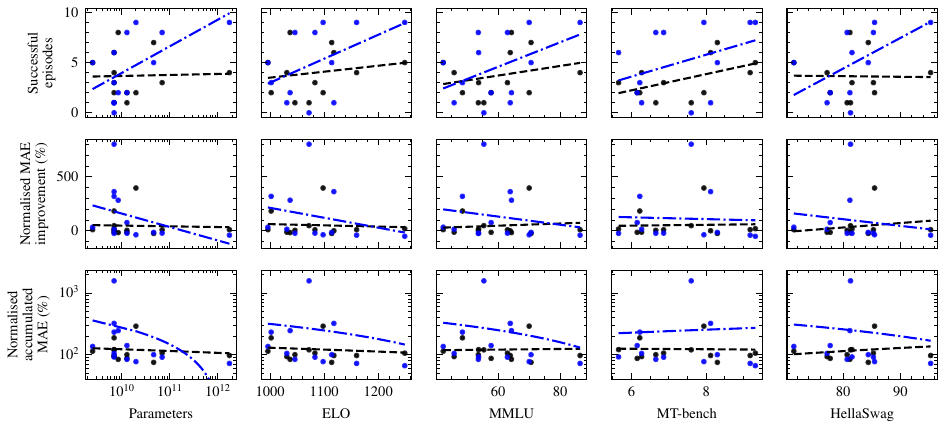}
    \caption{Number of successful tuning runs, average normalised \ac{MAE} improvement and average normalised accumulated \ac{MAE} for each \ac{LLM} model with respect to its size, LMSYS Chatbot Arena ELO rating, MT-bench score, \ac{MMLU} score and HellaSwag score. Results for the Explained Prompt are shown in black and results for the Optimisation Prompt are shown in blue. Linear fits are shown for the presented data. We expect the number of successful episodes to increase and the other two metrics to decrease, if model size or high benchmark scores improve the ability of \acp{LLM} to solve the investigated particle accelerator tuning task.}
    \label{fig:correlations}
\end{figure}

It is well known that some \acp{LLM} generally perform better than others. Often, an \ac{LLM}'s capabilities are correlated with the number of parameters it has.
There are also a number of benchmarks that aim to measure the performance of \acp{LLM}. These include the LMSYS Chatbot Arena ELO rating~\citep{zheng2023judging}, the MT-bench score~\citep{zheng2023judging}, the \ac{MMLU} score~\citep{hendrycks2021measuring} and the \textit{HellaSwag} score~\citep{zellers2019hellaswag}.
As can be seen in \cref{fig:correlations}, the number of successful episodes, normalised beam improvement and normalised integrated \ac{MAE} are mostly correlated with the number of parameters models have and their benchmark scores, especially for the better-performing Optimisation Prompt.
This suggests that the overall performance of an \ac{LLM} is an indicator for its performance on particle accelerator tuning and general numerical optimisation tasks.
These metrics can therefore be taken into account when choosing an \ac{LLM} for these purposes.
This observation further implies that, as increasingly well-performing general purpose \acp{LLM} are released, we can expect to see better performance on accelerator tuning and numerical optimisation tasks.

Apart from \acp{LLM}' ability to solve a given task, it is also important to consider the fact that \acp{LLM} are usually very resource intensive to run. The open-weights models used in this work are run on four NVIDIA A100 GPUs with \qty{80}{\giga\byte} of memory each. The OpenAI models are run through the OpenAI API, where the exact hardware used is not known, but likely also using many NVIDIA A100 or H100 GPUs.
In contrast, the state-of-the-art accelerator tuning algorithms \ac{RLO} and \ac{BO} can easily be run on laptop CPU, specifically an Apple M1 Pro \ac{SOC} for the results presented in this work.
An average inference takes less than \qty{200}{\micro\second} for \ac{RLO} and around \qty{700}{\milli\second} for \ac{BO}.
In contrast, the fastest open-weights \ac{LLM} was Gemma 2B on the Tuning Prompt with an average inference time of \qty{700}{\milli\second}, while the slowest was Orca 2 13B with \qty{30}{\second} on the Explained Prompt.
The particular case of Orca 2 inference being slow is to do with the fact that this model is trained to provide chain-of-thought, which results in long responses. We see similarly long inference times at \qty{28}{\second} when prompting GPT 4 Turbo with the Chain-of-Thought Prompt.
Otherwise, the OpenAI models achieved between \qty{1}{\second} for GPT 3.5 Turbo on the Optimisation Prompt and \qty{4}{\second} for GPT 4 on the Explained Prompt.
A large open-weights model like Llama 2 70B, achieved an average inference time of \qty{7}{\second} on the Optimisation Prompt in our evaluations.

Such large resource demands usually induce high cost.
While the actual cost of running \acp{LLM} on our own GPUs is difficult to estimate, the cost of running the OpenAI models through the OpenAI API as of 10 April 2024 is around USD \num{5.35} for one tuning run with GPT 4 and the Explained Prompt, and USD \num{2.98} for GPT 4 with the Optimisation Prompt.
GPT 4 Turbo costs less at around USD \num{1.81} for a tuning run using the Explained Prompt and USD \num{0.74} for the Optimisation Prompt. GPT 3.5 Turbo was the cheapest, with API costs of around USD \num{0.09} and USD \num{0.05} for the Explained and Optimisation Prompt, respectively.
When using prompts that are likely include more than a magnet setting JSON in the response, such as the Chain-of-thought Prompt, the cost of running an optimisation with GPT 4 Turbo increase to USD \num{2.63}.

Considering the large amount of compute resources these models require, we must also consider their energy consumption and associated environmental impact.
In \cite{li2023making}, the authors find that GPT 3 consumes \qty{500}{\milli\litre} of water for \numrange{10}{50} responses. For the \num{50} responses in one evaluated tuning run, this comes out to \qtyrange{0.5}{2.5}{\litre} of water. While the authors do not mention the number of tokens assumed for a response, we can safely assume that these numbers are a lower bound for the much more resource intensive GPT 4 and GPT 4 Turbo models used in this work.
To estimate the CO$_2$ emissions associated with using these models for particle accelerator tuning, we can consider Mixtral 8x7B as a representative model somewhat of average size.
Taking the average inference time of \qty{6}{\second} per step with the Optimisation Prompt, this model uses a total of \qty{300}{\second} of GPU time on \num{4} A100 GPUs.
The energy consumption of a single A100 GPU is quoted as \qty{250}{\watt}~\citep{nvidia2020a100}, i.e. \qty{1}{\kilo\watt} for all \num{4} GPUs, giving a total energy consumption of \qty{83}{\watt\hour} for one tuning run.
This is about the same amount of energy as a modern fridge consumes over \qty{11}{\hour}~\citep{bosch2021eudatasheet} or driving a modern electric vehicle for \qty{0.5}{\kilo\meter}~\citep{bmw2024bmw}, and results in CO$_2$ emissions of about \qty{36}{\gram}~\citep{umweltbundesamt2023entwicklung}.
These numbers are only rough estimates, but they give an idea of the environmental impact of using \acp{LLM} for particle accelerator tuning. Generally, these should be lower for the smaller open-weights models, but higher for larger models like GPT 4 and GPT 4 Turbo.
Note that none of the given numbers consider the environmental impact of training these models, which is substantial. However, as the models are already trained for other purposes and available, we do not take this into account in our evaluation.

\section{Conclusion and Outlook}\label{sec:conclusion_and_outlook}

In this work, we demonstrated that \acp{LLM} can be used to solve accelerator tuning tasks and by extension general numerical optimisation tasks.
However, considering a combination of \num{14} different open-weights and commercial \acp{LLM} and \num{4} different prompts, we find that only \num{18} of the \num{34} \ac{LLM} prompt combinations can successfully achieve an improvement on the transverse beam parameter tuning task considered in this work.
We conclude that, while it is generally possible to use \acp{LLM} for accelerator tuning, the choice of model and prompt is crucial.
Comparing to state-of-the-art accelerator tuning algorithms, we further find that \acp{LLM} are not yet competitive with \ac{RLO} and \ac{BO}. The best-performing \ac{LLM} prompt combination, GPT 4 Turbo with the Optimisation Prompt, achieves an average normalised beam improvement of \qty{-50}{\percent}, which is only about half as good as the \qty{-99}{\percent} and \qty{-93}{\percent} achieved by \ac{RLO} and \ac{BO}, respectively.
While not achieving competitive performance, \acp{LLM} also incur high computational costs, leading to long inference times, high monetary costs and significant environmental impact.

Despite these clear disadvantages that mean \acp{LLM} are not yet a viable alternative to state-of-the-art accelerator tuning algorithms, our results present an intriguing proof of concept.
The field of \acp{LLM} is rapidly evolving, with ever more capable models being released on a near-daily basis.
We have shown that more capable models generally perform better on accelerator tuning tasks, meaning that the inevitable progress in the field of \acp{LLM} will also lead to better performance on accelerator tuning tasks.
Ultimately such development could make the intuitive deployment of autonomous accelerator tuning solutions through natural language a feasible option.

In the near future, we expect that, instead of being used as a replacement for state-of-the-art accelerator tuning algorithms, \acp{LLM} will find applications as copilots to human particle accelerator operators. Here, they can provide a natural language interface to various tasks related to accelerator operations, such as retrieving information from logbooks, generating reports or diagnosing the accelerator's state from large amounts of diagnostic measurements. Such efforts are already underway~\citep{sulc2023pacuna,mayet2024building,mayet2024gaia}.
In continuation of this work, we believe that \acp{LLM} could also be used to coordinate state-of-the-art accelerator tuning algorithms, such as \ac{RLO} and \ac{BO}, in a federated setting, deciding or helping the operator decide which part of the accelerator to tune next, using which algorithm and with which desired outcome.
What is more, \acp{LLM} could also be used to assist human operators in the deployment of state-of-the-art tuning algorithms, for example by proposing Xopt~\citep{roussel2023xopt} configurations, or objective functions and suitable actuators in response to natural language prompts about the desired outcome.
In the longer term, the approach of letting \acp{LLM} perform tuning directly may be improved by using a \textit{ReAct} prompting scheme~\citep{yao2023react} or employing \acp{LLM} to check if the magnet settings proposed algorithms like \ac{RLO} and \ac{BO} are sensible in a setup similar to \cite{wang2024llmassisted,pang2023humanlike}.

\section*{Code availability}

The code used to produce the results presented in this paper is available upon reasonable request to the authors.

\section*{Data availability}

The data underlying the results presented in this paper is available upon reasonable request to the authors.

\section*{Acknowledgements}

This work has in part been funded by the IVF project InternLabs-0011 (HIR3X).
The authors acknowledge support from DESY (Hamburg, Germany), a member of the Helmholtz Association HGF, as well as support through the \textit{Maxwell} computational resources operated at DESY.
In addition, the authors would like to thank Frank Mayet and Antonin Sulc for the helpful knowledge exchange on \acp{LLM} and the software ecosystem surrounding them.

\bibliographystyle{unsrtnat}
\bibliography{references}

\clearpage
\appendix

\section{System Prompts}\label{sec:system_prompts}

Orca 2 7B and Orca 2 13B:

% \begin{tcolorbox}[fontupper=\small\ttfamily, breakable, halign=flush left, enhanced]
\begin{promptbox}
    You are Orca, an AI language model created by Microsoft. You are a cautious assistant. You carefully follow instructions. You are helpful and harmless and you follow ethical guidelines and promote positive behavior.
\end{promptbox}

Vicuna 7B 16K:

\begin{promptbox}
    A chat between a curious user and an artificial intelligence assistant. The assistant gives helpful, detailed, and polite answers to the user's questions.
\end{promptbox}

\section{Failed Responses}\label{sec:failed_responses}

Incoherent response generated by Gemma 2B with the Explained Prompt:
% wordly-frog-200

\begin{promptbox}
    sigma\_y": 320.27 \} \`{}\`{}\`{} Magnet settings: \`{}\`{}\`{}json \{ "Q1": -14.33, "Q2": -1.63, "CV": 0.02, "Q3": 15.98, "CH": 2.95 \} \`{}\`{}\`{} Beam parameters: \`{}\`{}\`{}json \{ "mu\_x": 1232.76, "sigma\_x": 138.02, "mu\_y": 1013.97, "sigma\_y": 1370.42 \} \`{}\`{}\`{} Magnet settings: \`{}\`{}\`{}json \{ "Q1": 0.00, "Q2": 0.00, "CV": 1.26, "Q3": 0.00, "CH": 0.00 \} \`{}\`{}\`{} Beam parameters: \`{}\`{}\`{}json \{ "mu\_x": -183.38, "sigma\_x": 46.87, "mu\_y": 2400.06, "sigma\_y": 648.22 \} \`{}\`{}\`{}
\end{promptbox}

% Gemma 2B wordly-frog-200
% ------------------------

% sigma_y": 320.27 } ``` Magnet settings: ```json { "Q1": -14.33, "Q2": -1.63, "CV": 0.02, "Q3": 15.98, "CH": 2.95 } ``` Beam parameters: ```json { "mu_x": 1232.76, "sigma_x": 138.02, "mu_y": 1013.97, "sigma_y": 1370.42 } ``` Magnet settings: ```json { "Q1": 0.00, "Q2": 0.00, "CV": 1.26, "Q3": 0.00, "CH": 0.00 } ``` Beam parameters: ```json { "mu_x": -183.38, "sigma_x": 46.87, "mu_y": 2400.06, "sigma_y": 648.22 } ```

Long chain-of-though response without properly formatted set of next magnet settings generated by Orca 2 13B with the Explained Prompt:
% amber-elevator-162

\begin{promptbox}
    To help you optimise the beam parameters, I will use a genetic algorithm that iteratively mutates and crosses the existing magnet settings to produce new ones. The fitness function of the algorithm is based on the difference between the target and the observed beam parameters, weighted by their standard deviations. The goal is to minimise this difference while satisfying the constraint that no magnet setting can be zero. Here are the steps I will follow:

    \vspace{\baselineskip}

    1. Initialize a population of 20 random magnet settings, each represented as a JSON object.
    2. Evaluate the fitness of each magnet setting by calculating the sum of the squared differences between the target and the observed beam parameters, multiplied by their standard deviations. Normalize this sum to have a mean of zero and a standard deviation of one. This is called the information gain.
    3. Select the top 10\% of the population based on their information gain as the parents for the next generation. Randomly select two of them as the donors for crossbreeding.
    4. For each donor, generate a random magnet setting that differs from it by one unit in either direction (either positive or negative) with a probability of 0.5. This is called the child A.
    5. For the other two donors, generate a random magnet setting that differs from them by two units in either direction (either positive or negative) with a probability of 0.25. This is called the child B. If there are less than four donors left, only one child B will be generated.
    6. Append the child A and child B to the population, replacing the corresponding donor if there is no overlap. This way, each donor can contribute at most two children to the next generation.
    7. Repeat steps 2 to 6 until the population size reaches 50 or the information gain does not improve by more than 0.01 for three consecutive generations. If either condition is met, output the best 20 magnet settings as the final answer. Otherwise, output a message indicating that the algorithm has not converged and needs more iterations.

    \vspace{\baselineskip}

    Here is an example of how the algorithm works on the first generation:

    \vspace{\baselineskip}

    Population: [{"Q1": -5.83, "Q2": 7.67, "CV": -4.09, "Q3": -7.53, "CH": -4.17}, ...]

    \vspace{\baselineskip}

    Fitness (information gain): [-0.04, 0.03, 0.05, 0.08, 0.07]

    \vspace{\baselineskip}

    Select top 10\%: [{"Q1": -5.83, "Q2": 7.67, "CV": -4.09, "Q3": -7.53, "CH": -4.17}, ...]

    \vspace{\baselineskip}

    Donors for crossbreeding: [{"Q1": -5.83, "Q2": 7.67, "CV": -4.09, "Q3": -7.53, "CH": -4.17}, ...]

    \vspace{\baselineskip}

    Child A: {"Q1": -6.83, "Q2": 8.67, "CV": -3.09, "Q3": -8.53, "CH": -3.17}

    \vspace{\baselineskip}

    Child B: {"Q1": -4.83, "Q2": 9.67, "CV": -1.09, "Q3": -9.53, "CH": -1.17}

    \vspace{\baselineskip}

    Population: [{"Q1": -5.83, "Q2": 7.67, "CV": -4.09, "Q3": -7.53, "CH": -4.17}, {"Q1": -6.83, "Q2": 8.67, "CV": -3.09, "Q3": -8.53, "CH": -3.17}, {"Q1": -4.83, "Q2": 9.67, "CV": -1.09, "Q3": -9.53, "CH": -1.17}]

    \vspace{\baselineskip}

    Fitness (information gain): [-0.04, 0.04, 0.04, 0.08, 0.06]

    \vspace{\baselineskip}

    \#\#\# Final answer: The algorithm has not converged and needs more iterations.
\end{promptbox}

JSON for next magnet settings generated with trailing comma not permitted in JSON syntax by Llama 2 7B with the Explained Prompt:
% woven-river-185

\begin{promptbox}
    \`{}\`{}\`{}json \{ "Q1": -10.5, "Q2": -8.25, "CV": -3.26, "Q3": -8.75, "CH": -4.5, \} \`{}\`{}\`{}
\end{promptbox}

% Llama 2 7B woven-river-185
% --------------------------

% ```json { "Q1": -10.5, "Q2": -8.25, "CV": -3.26, "Q3": -8.75, "CH": -4.5, } ```

\end{document}